# A Spiking Neural Network Learning Markov Chain

**Mikhail Kiselev**
Chuvash State University
Cheboxary, Russia
`mkiselev@chuvsu.ru`

September 20, 2022

## Abstract

In this paper, the question how spiking neural network (SNN) learns and fixes in its internal structures a model of external world dynamics is explored. This question is important for implementation of the model-based reinforcement learning (RL), the realistic RL regime where the decisions made by SNN and their evaluation in terms of reward/punishment signals may be separated by significant time interval and sequence of intermediate evaluation-neutral world states. In the present work, I formalize world dynamics as a Markov chain with unknown a priori state transition probabilities, which should be learnt by the network. To make this problem formulation more realistic, I solve it in continuous time, so that duration of every state in the Markov chain may be different and is unknown. It is demonstrated how this task can be accomplished by an SNN with specially designed structure and local synaptic plasticity rules. As an example, we show how this network motif works in the simple but non-trivial world where a ball moves inside a square box and bounces from its walls with a random new direction and velocity.

## 1 Introduction

Reinforcement learning (RL) [1] is a key technique for creation of adaptive systems, which are trained to reach specified goals interacting with dynamic real-world environment (e. g. in robotics). At the same time, many researchers consider spiking neural networks (SNN) as a promising logical basis for this kind of intelligent systems because they add in a natural way the time dimension to computational power of traditional neural networks. It makes RL implementation in SNN very important problem.

There are two approaches to RL: model-free and model-based. The former is much simpler because it implies the "stimulus-reaction" operation mode – it is assumed that there is certain optimum action for each current world state and the network should learn it. However, the majority of real RL tasks do not fit this scheme because choice of the optimum decision has to be based on the previous history and only a small part of possible world states are explicitly indicated as desirable or undesirable by reward/punishment signals. However, the more realistic model-based RL is much harder to implement because it requires creation inside the network of an internal model of the world dynamics and its responses to actions of the network. If such a model is created, the network can use it to plan sequence of actions leading to obtaining the maximum reward.

In this paper, we consider an SNN structure, which partially solves this problem. Namely, we treat world dynamics as Markov chain [2]. It means that at every moment the world is in one of a finite set of states and that transitions between these states are stochastic and described by probabilities unknown a priori but depending only on the current and next states. We do not consider here the state transitions explicitly caused by network actions. Despite this, the problem remains actual because 1) it is frequent situation that network actions are rare and world dynamics are governed by its internal laws; 2) on the early stages of learning, network actions are almost random and therefore can be considered just as stochastic factors contributing to world's internal stochasticity; 3) even when changing network behavior changes significantly the transition probabilities, the Markov chain transition probability model learnt allows to roughly estimate proximity of the given world state to desirable or undesirable states. The SNN capable of learning the fully functional world model incorporating network actions in the explicit form will be a goal for the next stages of our research project.

I would like to stress on an important distinctive feature of the present study making the problem solved more realistic but much harder. The classic definition of Markov chain does not include time explicitly (the discrete-time Markov chain). Namely, it is assumed that there are discrete time steps $i$ and the system resides in a certain state $s_i$ during the step $i$ after which moves to some other state $s_{i+1}$ (however, the reflexive transitions $s_{i+1} = s_i$ are also possible). However, the real RL tasks "live" in our continuous time. It means that the system can reside in any state during some indefinite time and these time intervals may be different for the same state. As we will see, our SNN can tackle this problem and, therefore, is capable of working in the real world. Another distinctive feature is usage of simple deterministic neurons easily implementable on the modern neuromorphic hardware (like Intel's Loihi [3]). Although there has been several implementations of SNNs learning Markov chains (see below) but I am unaware of existing implementations possessing these features.

SNNs learning Markov chains were explored mainly in the works devoted to the model-based RL. However, in contrast with model-free RL [4 - 9], works on implementation of model-based RL in SNN are few. As an example, [10] can be mentioned. However, a significant part of the learning system described in [10] (the arbitration component) is implemented outside the network – so that it cannot be called a purely SNN solution. Besides that, the SNN operation is controlled by an external process (for example, lateral inhibition is implemented by external "winner-takes-all" mechanism instead of inhibitory connections) and simulation time is discrete. The same is true for the model-based RL system including SNN, which is considered in [11]. Mechanism of world model acquisition there also includes the significant non-neural components. Another model-based RL implementation in a recurrent SNN is described in [12]. However, in this work, the world model is not learnt – instead, it is hardwired in the network in the form of the lateral weights $w_{ki}$. At the time of writing this article, I am aware of only one SNN implementation of model-based RL where the world model is learnt. It is described in [13]. World model learning in this study has the form of adjusting weights of certain interneuron connections but rules for this adjustment are not local – they are not formulated in terms of pre-/post-synaptic neuron spike timing and other properties of the neurons connected by the modified synapse, as it is required for a genuine SNN-based solution.

In several papers (e. g. [14 - 16]), the close question is considered: how current state of hidden Markov model (HMM) can be inferred by an SNN (HMM inference). In these papers, it is discussed how information necessary for this inference is represented in SNN but the problem of learning the state transition probabilities is not solved.

Another similar but significantly distinct problem studied in many works is sequence learning. Sequence learning is the network ability to learn reproducing sequences of values encoded in some form (sometimes together with time intervals between these values) which were presented to the network in the training phase. Many approaches to solution of this problem by SNN have been proposed [17 - 20]. In this works, the goal is to reproduce some finite set of possible deterministic sequences while, in our case, the set of possible sequences is infinite and the sequences are stochastic.

At last, it should be noted that traditional (non-spiking) neural networks learning Markov chain have been described in several papers (for example, [21]).

As it was said, in contrast with the above-mentioned works, I use only SNN structures to learn world models in the form of Markov chain – without any extra-network mechanisms, and using only local synaptic plasticity rules.

I would like to add that the following important problem has not been considered in the existing works. In all implementations of model-based RL in SNN (like [10 - 13]), the world model has the similar form: every neuron in a certain neuron population represents one world state and weights of connections between these neurons correspond to the probabilities of transitions between the corresponding states. Obviously, in realistic complex tasks, the count of possible world states may be huge. Of course, if the set of possible transitions from any given state is known and not large then the weight matrix is sparse, every state neuron has moderate number of synapses, and it makes no problem. However, if the world model (= the weight matrix) is not known and should be learnt then every state neuron should have as many synapses as the number of possible world states. It is unrealistic. Even in the brain, there are tens billions neurons but every neuron has not more than few tens of thousands synapses. This limit is even stricter in the modern neuroprocessors – for example, the maximum number of synapses of neurons implemented in the TrueNorth neurochip [22] is only 512.

In the next Section, we discuss how this and other problems of Markov chain learning are solved in my approach with help of the specially designed network architecture and the simple synaptic plasticity rules.

## 2 Materials and Methods

### 2.1 SNN Architecture

As it was said, in SNN implementations of RL, the world states correspond to certain neurons, while the world model has the form of the matrix **W** containing weights of links between these neurons. I use the similar approach but instead of a single neuron, every world state is represent by a special neuronal structure, which will be referred to as *column* (here, I do not claim any direct analogy with cortical columns – this term is rather a metaphor; nevertheless some analogy may be possible). Respectively, the transition probabilities are represented as weights of certain connections between neurons belonging to different columns. Every column has its own input node. Spikes coming to input node of a column indicate the respective current world state. More precisely, to indicate a state, the spikes should come with the inter-spike interval (ISI) not longer than $t_{ISI}$ (it is a constant of our model fixing its time scale). In this case, we say that the given column is active. It is assumed that only one column can be active at any moment of time.

This columnar network structure is depicted on Figures 1 and 2 showing 3 levels of structural hierarchy of the SNN. After the general consideration of its components, it will be described formally. Figure 1 shows structure of a single column (all columns in this network have the same structure).

As it was said, the sources of signals for the whole network are input nodes indicating current world state – one node per column. The input node stimulates the neuron denoted as MEM on Figure 1. This stimulation is so strong that forces MEM to fire unconditionally. The MEM neurons have memory property. Once starting to fire, they emit spikes periodically for indefinite time until their firing is stopped by an inhibitory spike. This feature can be easily implemented using reflexive excitatory connection or, say, by a pair of neurons forming the loop. To simplify this construction, I included the memory property immediately in the neuron model. Such a *memory neuron,* being forced to fire, begins emitting spikes with the period $\tau_M$ until it receives a spike on its inhibitory synapse. The purpose of the MEM neuron is to store information about the previous world state. It is necessary to accumulate statistics of state transitions. However, in order to collect this statistics, we should distinguish between the current and most recent world states. This is the purpose of the GATE neuron. It receives strong inhibition from the input node so that it cannot fire while its input is active. But after this inhibition ceases this neuron begins to fire as a result of stimulation from MEM (which continues to fire even after end of the input node activity). Therefore, activity of GATE indicates that the corresponding world state was observed recently but now it is not the current state. It is important that the GATE neurons have dynamic membrane potential threshold growing with each firing and slowly relaxing to its basic level. It means that the GATE neurons are very active at the beginning of a new state. In this time, they replicate the frequent MEM firing. But afterwards, they begin to fire rarely because their threshold potential increases. Thus, it can be said that high activity of GATE indicates moments of state transitions.

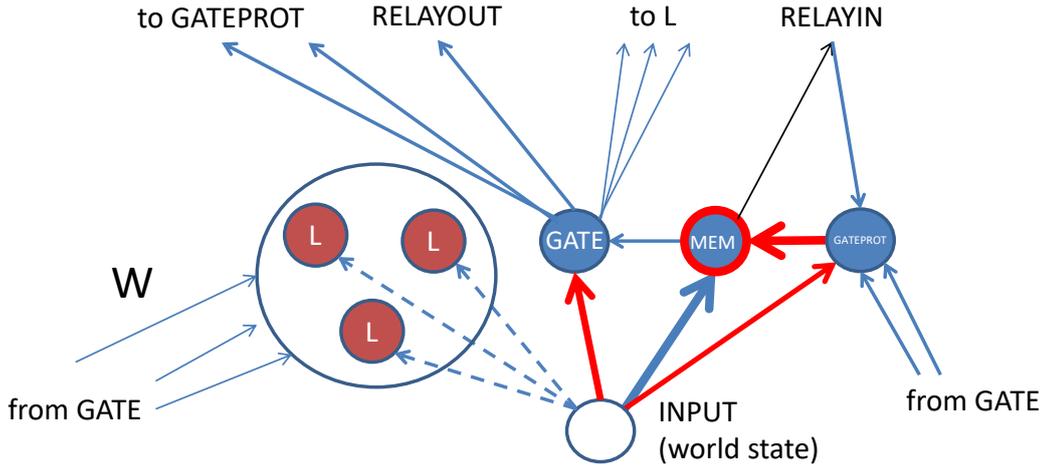

Figure 1. The schematic structure of a single SNN column. The arrow color codes connection type: blue - excitatory; red – inhibitory. The dashed arrows denote plasticity modulating ("dopamine") connections. Line thickness corresponds to connection strength. Thick red circle means memory property. See the detailed description in the text.

Obviously, this mechanism lacks an important component, which should suppress activity of columns corresponding to the world states from distant past. Since we restrict ourselves to Markov model, all the past world states except the most recent one are not relevant and should be ignored. This task is accomplished by the inhibitory GATEPROT neuron, which terminates the infinite spike train generated by MEM by its inhibitory spike. This neuron is unconditionally inhibited by activity of the input node. Therefore, it cannot block the MEM neuron whose column is active. But when the column becomes inactive, GATEPROT is not blocked anymore. However, in order to fire, the GATEPROT neuron should receive some stimulation. It receives it from GATE neurons of the other columns (excluding the column of its own). This configuration solves the problem. Indeed, since activity of a GATE neuron indicates the most recent (but not current) world state, it should stop activities of all other "previous" columns. It is done by sending by the GATE neuron the powerful stimulation to all the GATEPROT neurons except the GATEPROT neuron of its own column. As it was said above, the MEM neuron of the active column also remains unaffected by this global inhibition caused by the GATEPROT neurons.

From all these considerations, we see that at any moment we know the current world state (it is indicated by input node of some column) and the previous world state (it is indicated by activity of the GATE neuron of another column). Therefore, we can gather statistics of state transitions to learn the world dynamics model in the form of Markov chain. It is done by the learning neurons labelled by the 'L' letter on Figure 1. Every L neuron has two types of synaptic connections: from the GATE neurons of the other columns and from the input node of its own column. The former inputs are excitatory and plastic; the latter input plays the role of plasticity modulator. The plasticity rules are simple. Whenever the neuron receives a spike via its plasticity inducing synapse, its plastic synapses having received spikes during the time $\tau_P$ in the recent past are potentiated by the same constant value $A$. The second rule states that the total synaptic strength of a neuron should remain constant. Therefore, all the other plastic synapses are depressed by the same value $n_S^+ A/(n_S - n_S^+)$, where $n_S^+$ is the number of plastic synapses having received spikes, $n_S$ – the total number of plastic synapses. It is evident that after sufficiently long learning time, weights of these connections (they are labelled by 'W" on Figure 1) will depend monotonously on the respective state transition probabilities. The weight matrix **W** can be used to estimate the ratio of probabilities of transitions between the current state and the other states and therefore it can be called the world model if the world state transitions is considered as Markov chain. It is important that duration of any state has little effect on **W** modification – **W** is updated mainly when the GATE neurons are most active i. e. just after state transition.

In principle, this structure is sufficient to learn a simple world model, with small number of possible states. However, as it was said in Introduction, in the real-world tasks we will face the hard problem of numerous world states that would require providing L and GATEPROT neurons with huge amount of synapses. Below, I describe my approach to solution of this problem.

For L neurons and the case when the network topology cannot be changed dynamically (if synapses cannot be created or removed during network functioning) the only solution seems to be increasing number of L neurons inside one column (it is why the group of L neurons is shown on Figure 1). If there are N L neurons inside one column then each L neuron can have N times less synapses provided that sets of presynaptic neurons of these L neurons do not intersect.

For GATEPROT neurons, the solution can be more economic since their connections are not plastic and not specific. It is achieved due to adding special relay neurons shown as RELAYOUT and RELAYIN neurons on Figure 2. The scheme works in the following way. The columns are united in the groups (one of these groups is shown on Figure 2A). Every groups has one RELAYOUT neuron and one RELAYIN neuron. The RELAYOUT neuron is unconditionally excited by any GATE neuron in its group. It broadcasts its excitatory spikes to RELAYIN neurons in all other groups. This excitation is subthreshold – the spikes from RELAYOUT by themselves cannot force RELAYIN to fire. To do it, the RELAYIN neuron should receive additional excitation from some MEM neuron in its group. If it does then it fires and activates GATEPROT neurons in its groups so that these neurons inhibit MEM neurons in all inactive columns. In such a way, the goal is reached – the GATE neuron in the column corresponding to the previous state suppresses MEM neurons in all other inactive columns network-wide. But due to the hierarchical organization, it is reached at a price of much less number of synapses. It is equal to the number of columns in the group for the GATEPROT and RELAYOUT neurons and to the sum of the number of groups and the number of columns in one group – for RELAYIN neurons.

## 2.2 Neuron Model

Now let us consider the formal model of neuron and synaptic plasticity. I used a slightly generalized version of the simple but functionally rich LIFAT (leaky integrate-and-fire neuron with adaptive threshold) neuron model [23]. It can be efficiently implemented on the modern neurochips (e.g. Loihi). The simplest current-based delta synapse model was used for all synapses. Every time the synapse receives a spike, it instantly changes the membrane potential by the value of its synaptic weight, which may be positive or negative depending on the synapse type. The neuron state at any moment $t$ is described by its membrane potential $u(t)$ and its threshold potential $u_{THR}(t)$. Dynamics of these values are defined by the equations

$$\begin{cases} \frac{du}{dt} = -\frac{u}{\tau_v} + \sum_{i,j} w_i^+ \delta(t - t_{ij}^+) - \sum_{i,j} w_i^- \delta(t - t_{ij}^-) \\ \frac{du_{THR}}{dt} = -a\,\text{sgn}(u_{THR} - 1) + \sum_k \widehat{T} \delta(t - \hat{t}_k) \end{cases} \quad (1)$$

and the conditions that $u$ is hard limited from below by the value $u_{MIN}$ and that if $u$ exceeds $u_{THR}$ then the neuron fires and value of $u$ is reset to 0. All potentials are rescaled so that after the long absence of presynaptic spikes $u \to 0$ and $u_{THR} \to 1$. The meaning of the other symbols in (1) is the following: $\tau_v$ – the membrane leakage time constant; $a$ – the speed of decreasing $u_{THR}$ to its base value 1; $w_i^+$ - the weight of $i$-th excitatory synapse; $w_i^-$ - the weight of $i$-th inhibitory synapse; $t_{ij}^+$ - the time moment when $i$-th excitatory synapse received $j$-th spike; $t_{ij}^-$ - the time moment when $i$-th inhibitory synapse received $j$-th spike; $\widehat{T}$ – $u_{THR}$ is incremented by this value when the neuron fires at the moment $\hat{t}_k$. It should be noted that only the GATE neurons are LIFAT neurons, all other neurons are LIF ($\widehat{T} = 0$).

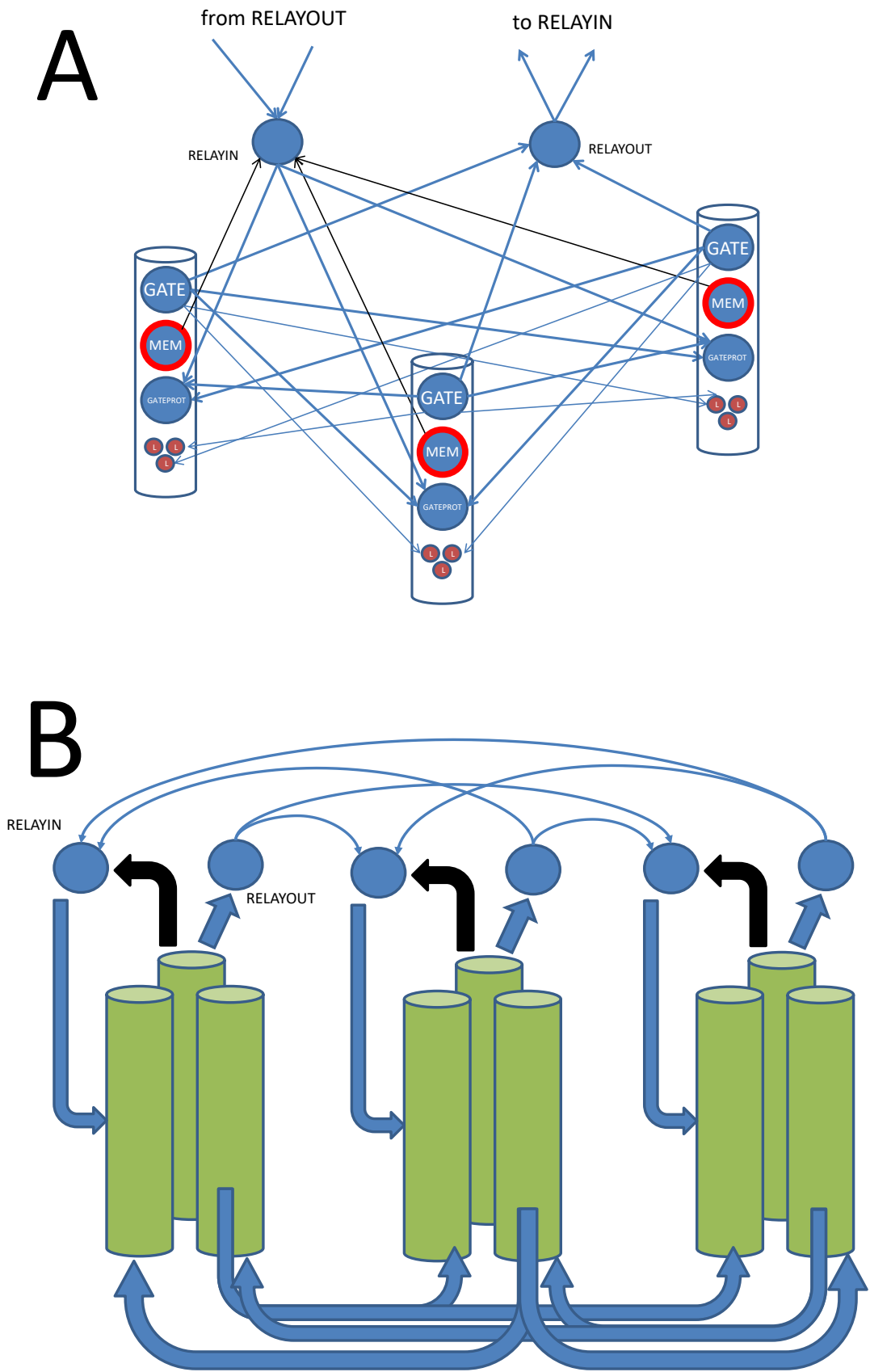

Figure 2. The hierarchical structure used to decrease number of synapses of the GATEPROT neurons. A. The connections inside one column group. Inter-group connections of the L neurons are not shown. B. The whole network including several column groups. The thick arrows at the bottom are connections of the L neurons. See the detailed description in the text.

Besides that, the memory property can be added to this model. Neuron has the parameter called memory spike train period $\tau_M$. If this parameter is defined (not equal to infinity), then after every firing, the neuron internal timer is reset to the value $\tau_M$. When this timer reaches zero value the neurons fires (independently of its current membrane potential) and the timer is reset again immediately. It is equivalent to presence of a very strong reflexive connection with delay time equal to $\tau_M$. This infinite periodical firing is terminated by a spike obtained by any neuron's inhibitory synapse. Only the MEM neurons have this property in the SNN described.

The synaptic plasticity rules are local and include only characteristics of activity of the pre- and post-synaptic neurons. Its main distinctive feature is the same as in our previous research works [24 - 26]. Namely, the synaptic plasticity rules are applied to the so called *synaptic resource W* instead of the synaptic weight *w*. There is functional dependence between *W* and *w* expressed by the formula

$$w = w_{min} + \frac{(w_{max} - w_{min})\max(W, 0)}{w_{max} - w_{min} + \max(W, 0)} \qquad (2)$$

where $w_{min}$ and $w_{max}$ are constant (in this study, $w_{min} = 0$).

In this model, the weight values lay inside the range $[w_{min}, w_{max})$ - while *W* runs from $-\infty$ to $+\infty$, *w* runs from $w_{min}$ to $w_{max}$. In the SNN described only connections between the GATE and L neurons are plastic.

The L neurons have a special plasticity inducing synapse. When it obtains spike, synaptic resources of all plastic synapses of the neuron having received at least one presynaptic spike during the time interval of the length $\tau_P$ in the past are modified by the value equal to the weight of this plasticity inducing synapse.

In order to introduce competition between the synapses, one more rule - constancy of neuron's total synaptic resource was added. Without it, all synapses would eventually reach saturation. To implement this rule, the additional component of neuron state *b* is introduced. Whenever some synaptic weight is increased by the value $\Delta$, *b* is incremented by the same value $\Delta$. When the value of *b* becomes sufficiently great, the weight renormalization procedure is evoked – all weights are decreased by the value *b* divided by the number of plastic synapses. After that, *b* is reset to 0.

## 2.3 Specifications for Input Signal and Neuron Model

The architecture described in this article is thought to be universal, unbound to any concrete task. However, it can work only if certain general requirements for input signal are satisfied. At first, I will specify these requirements and give the exact values of parameters of all types of neurons constituting the network, then I will describe a simple but non-trivial task used to test my approach. At last (in the next Section), the results obtained by this network in this task will be discussed.

The simulation of the network functioning was performed in discrete time – one time step corresponds to one network state update. Like in many similar works on computer simulation of SNN, we assume that one time step equals to 1 msec.

The requirements to the input signal were partially formulated above. It is required that only one input would be active at any moment of time. "Active" means that the node emits spikes with the ISI not greater than $t_{ISI}$. I set this value equal to 10 msec – this value fixes the scale for all other time constants. Simultaneous activity of several inputs will cause unpredictable update of the **W** matrix. The ISI value greater than $t_{ISI}$ is interpreted as exiting the current state.

The neuron parameters are the following (only the values essential from point of view of correct SNN operation are considered):

- **For MEM neurons.** $\tau_M = 3$ msec. It should be significantly less than $t_{ISI}$ for precise determination of state transition moment. The excitatory synapse weight equals to 10 – it should cause firing in any case – even in case of inhibition from GATEPROT. The inhibitory weight should be less (4).
- **For GATE neurons.** $\tau_v = 10$ msec, $u_{MIN} = -5$, $a = 0.01$ msec$^{-1}$, $\hat{T} = 0.1$, the inhibitory synapse weight equals to 6, the excitatory synapse weight equals to 1.15. These values are tuned to satisfy two conditions: 1) inhibition from the input node should prevent GATE firing at least during the time $t_{ISI}$ after every input spike; 2) after exiting the current state (when inhibition from the input nodes weakens) GATE should fire frequently (repeating MEM 3 msec firing) but then it should decrease firing frequency reacting to each 3$^{rd}$ spike from MEM. The 2$^{nd}$ condition is necessary because at first new active GATE neuron should suppress activity of the other GATE neurons with help of strong stimulation of the GATEPROT neurons, but after that this stimulation should be lowered thus making possible blocking the GATEPROT neurons in the new currently active column. Besides that, it binds the weight modification process to state transitions.

- **For GATEPROT neurons.** $\tau_v$ = 10 msec, $u_{MIN}$ = -3, the inhibitory synapse weight equals to 5, the weight of all excitatory synapses equals to 1.15. Again, these values is the result of tuning targeted at the correct state sequence detection. After exiting the current state (after time $t_{ISI}$ since the last spike from the input node), the recently active column becomes "previous" column. This status of this column should last at least $t_{ISI}$. It should be guaranteed that during this time interval the GATEPROT neuron remains inhibited by the input node. But after this time interval, it should be possible to activate it by 1 incoming spike from some GATE neuron. The improper setting of the parameters of the GATE and GATEPROT neurons may lead, for example, to the situation of the undesired MEM inhibition when in case of the fast state transitions X→Y→Z, still lasting activity of X GATE could suppress Y MEM thus preventing fixation of the Y→Z transition.
- **For L neuron.** $\tau_P = t_{ISI}$, which is quite obvious choice. The plasticity inducing synapse weight determines learning speed. It can be set equal to some small value like 0.1. It is the only parameter of my model, which may require tuning to the concrete task.
- **For RELAYOUT neuron.** All synaptic weights are set to 1.15 – just to force the neuron to fire every time it receives a spike.
- **For RELAYIN neuron.** All synaptic weights are set to 0.7 – to activate the neuron by 2 spikes – from RELAYOUT and from MEM.

Now, the SNN is described completely – let us consider an example of its application.

## 2.3 The Test Markov Chain Learning Task – Ball Moving inside Square Box

In order to test capability of the SNN proposed of learning the world dynamics treated as a Markov chain, I selected the following simple but non-trivial task. The ball moves in the square area of size 1×1 m without friction. When it hits a wall it randomly changes its movement direction (but it continues moving inside the box) and velocity (in the range [2.5, 10] m/sec). Thus, at every time moment the world state is described by the 4 values: (*x, y*) ball coordinates and ($v_x, v_y$) components of ball velocity. To make the world state discrete, I break [0, 1] range of coordinate values into 10 equal intervals of length 0.1; values of velocity components are binned into 5 intervals each so that the probabilities that the box velocity component measured at a random time moment is inside each interval are approximately equal. Therefore, the set of world states includes 2500 quaternions of discrete values (*x, y, $v_x$, $v_y$*) and transitions between them can be considered as Markov chain. It is obvious also, that transitions between states are not deterministic even when the ball is far from the walls – depending on the exact ball position and velocity, the state (2, 5, 2, 4) may be followed not only by the most probable state (2, 6, 2, 4) but also by (3, 5, 2, 4) and (1, 5, 2, 4). The state transition rule near the walls is completely different and even more stochastic. Thus, we see that this problem is sufficiently realistic and non-trivial.

Since there are 2500 possible world states, the SNN includes 2500 columns. The current state is indicated by activity of the respective input node emitting spikes with random ISI ranging from 3 to 5 msec.

To test scalability scheme described in Subsection 2.1, I introduced an artificial limitation that number of synapses should not exceed 1024 per neuron. It is necessary to include 3 L neurons in every column to satisfy this requirement. The columns were broken to 50 groups.

# 3 Results

All SNN emulation experiments took 1000 sec (1 000 000 time steps). This corresponds to approximately 20 000 – 30 000 state transitions.

The networks learns the world model in the form of synaptic weights. While these weights do not equal to transition probabilities, they monotonously depend on them. For this reason, Spearman rank correlation coefficient was used as an appropriate tool for assessment of quality of the world model learnt by the network,. Namely, the assessment procedure was the following.

The very long (100 000 sec) "ball in the box" emulation was run to calculate the accurate etalon state transition probabilities from the state transition counts. To obtain these counts, the ball coordinates and velocities were sampled periodically. Since by design of my SNN, it cannot detect world state if it lasts less than $t_{ISI}$ msec, the sampling period length should be few times greater than $t_{ISI}$. I selected it equal to 30 msec. If in two subsequent samples, ball states are different, the corresponding state transition counter is incremented.

The same procedure was repeated for a shorter emulation time (of length equal to the SNN emulation time – 1000 sec). It was done 30 times – for different ball trajectories. For each ball state, the set of states with non-zero number of transitions to this state was determined. Spearman correlation coefficient was calculated for counts of transitions to each state in the long and short emulations. Mean and standard deviation of these correlation coefficients were calculated and used to compare with the similar values obtained from world models learnt by the network. The mean value of correlation coefficient was determined to be equal to 0.59, standard deviation – to 0.34.

For the SNN world model, the synaptic weights **W** were used for calculation of Spearman coefficients. Only the state transitions, for which the synaptic weights were positive, were selected for this procedure. For each state, the set of states, transitions from which to this state were observed in the long simulation, or corresponding to positive synaptic weights, was determined. This time, the correlations between the state transition counts to the given state observed in the long simulation and the synaptic weights corresponding to these transitions in the trained SNN were calculated. The correlation coefficients were calculated for every ball state. The same procedure was repeated also 30 times for different ball trajectories – 30 different SNNs were trained. Again, the mean and standard deviation were calculated for all these coefficients. They were found to be 0.51 and 0.32, respectively.

We see that correlation of the synaptic weights W with true transition probabilities is close (inside the range of statistical fluctuations) to the maximum possible one. This result confirms efficiency of the considered SNN in tasks of learning world models in the form of Markov chain.

The C++ code implementing the 'ball in the box" world can be found at https://github.com/MikeKis/AShWin (the branch `RLv1optimized`, the project `LightSpotPassive`). The code calculating Spearman coefficients reported above is at the same place, the project `for_arxiv`.

# 4 Conclusion.

In this paper, we considered the SNN solving problem of learning the world model in the form of Markov chain under the realistic conditions:

- The signal indicating the current world state has the spiking form.
- World's dynamics unfold in continuous time, duration of each state is unknown a priori and may be different even for the same state.
- The number of different world states may be very great.

The solution of this problem is achieved due to the specially designed columnar network structure and tuned parameters of the neurons and the synapses. However, this fine-tuning is not based on specific properties of the concrete task (except the very general requirements to the input signal formulated in Subsection 2.3) and, therefore, this approach should work in wide range of tasks.

As an example, it was demonstrated how the network learns dynamics of a simple but non-trivial system – a ball moving chaotically inside a square box. It was shown using Spearman rank correlation coefficient, that synaptic weights of the plastic connections inside the SNN trained depend monotonously on the respective state transition probabilities.

This work is important for development of SNN implementation of model-based RL. This is the first step to automated creation by the network of a world model which could be used for planning actions leading to obtaining maximum reward. The next step in this direction, namely, creation of models, which could take into account impact of network decisions on its environment, should be a goal of the next stage of our research.

## Acknowledgements.

The present work is a part of the research project in the field of SNN carried out by Chuvash State University in cooperation with Kaspersky and the private company Cifrum.

Cifrum's GPU cluster was used for running the model verification procedure based on Spearman coefficients described in Section 3. All other computations were performed on my own computers equipped with GPUs. My own SNN emulation package called ArNI-X was used to obtain all results reported in this paper.

I thank communities of the Telegram groups "AGIRussia" and "NrmAI" for valuable discussion.

## References.

[1] R. Sutton and A. Barto, "Reinforcement Learning: An Introduction," A Bradford Book, 2014.

[2] P. A. Gagniuc, "Markov Chains: From Theory to Implementation and Experimentation," John Wiley & Sons, USA, 2017.

[3] M. Davies et al., "Loihi: A Neuromorphic Manycore Processor with On-Chip Learning," IEEE Micro, v 38, n 1, pp 82-99, 2018.

[4] R. Florian, "Reinforcement Learning through Modulation of Spike-Timing-Dependent Synaptic Plasticity", Neural Computation, v. 19, pp. 1468–1502, 2007.

[5] Ph. Weidel, R. Duarte and A. Morrison, "Unsupervised Learning and Clustered Connectivity Enhance Reinforcement Learning in Spiking Neural Networks" Frontiers in Computational Neuroscience, v 15, 2021 https://www.frontiersin.org/article/10.3389/fncom.2021.543872

[6] J. Jordan, Ph. Weidel and A. Morrison, "Closing the loop between neural network simulators and the OpenAI Gym," arXiv preprint arXiv:1709.05650, 2017.

[7] J. Jitsev, A. Morrison and M. Tittgemeyer, "Learning from positive and negative rewards in a spiking neural network model of basal ganglia," 2012 International Joint Conference on Neural Networks (IJCNN), pp. 1-8, 2012, doi: 10.1109/IJCNN.2012.6252834.

[8] T. Nakano, M. Otsuka, J. Yoshimoto, K. Doya, "A Spiking Neural Network Model of Model-Free Reinforcement Learning with High-Dimensional Sensory Input and Perceptual Ambiguity," PLoS ONE, 10, 3, e0115620, 2015, https://doi.org/10.1371/journal.pone.0115620

[9] D. Rasmussen, A. Voelker, C. Eliasmith, "A neural model of hierarchical reinforcement learning," PLoS ONE, 12, 7, e0180234, 2017, https://doi.org/10.1371/journal.pone.0180234

[10] R. Basanisi, A. Brovelli, E. Cartoni, G. Baldassarre, "A generative spiking neural-network model of goal-directed behaviour and one-step planning," PLoS Comput Biol, 16, 12, e1007579, 2020, https://doi.org/10.1371/journal.pcbi.1007579

[11] M. M. Shein, "A spiking neural network of state transition probabilities in model-based reinforcement learning," Master's thesis. University of Waterloo, 2017, https://uwspace.uwaterloo.ca/bitstream/handle/10012/12574/Shein_Mariah.pdf?sequence=3&isAllowed=y

[12] E. Rueckert, D. Kappel, D, Tanneberg, et al., "Recurrent Spiking Networks Solve Planning Tasks" Sci Rep 6, 21142, 2016, https://doi.org/10.1038/srep21142

[13] J. Friedrich, M. Lengyel, "Goal-Directed Decision Making with Spiking Neurons," Journal of Neuroscience. 36, 5, pp 1529–1546, 2016, https://doi.org/10.1523/JNEUROSCI.2854-15.2016

[14] S. Deneve, "Bayesian inference in spiking neurons," Advances in Neural Information Processing Systems, vol. 17, pp. 353–360, 2005.

[15] A. Tavanaei, A. S. Maida, "Training a Hidden Markov Model with a Bayesian Spiking Neural Network," J Sign Process Syst, 90, pp 211–220, 2018, https://doi.org/10.1007/s11265-016-1153-2

[16] Y. Zheng, Sh. Jia, Zh. Yu, T. Huang, J. K. Liu, Y, Tian, "Probabilistic inference of binary Markov random fields in spiking neural networks through mean-field approximation," Neural Networks, 126, pp 42-51, 2020, https://doi.org/10.1016/j.neunet.2020.03.003.

[17] Q. Liang, Y. Zeng, and B. Xu, "Temporal-Sequential Learning With a Brain-Inspired Spiking Neural Network and Its Application to Musical Memory," Front. Comput. Neurosci. 14, 51, 2020, doi: 10.3389/fncom.2020.00051

[18] Z. Liu, Th. Chotibut, Ch. Hillar, Sh. Lin, "Biologically Plausible Sequence Learning with Spiking Neural Networks," Proceedings of AAAI-20, pp 1316-1323, 2020.

[19] F. Ponulak, A. Kasiński, "Supervised learning in spiking neural networks with ReSuMe: sequence learning, classification, and spike shifting," Neural Comput., 22, 2, pp 467-510, 2010 doi: 10.1162/neco.2009.11-08-901

[20] J. Brea, W. Senn, and J.-P. Pfister, "Sequence learning with hidden units in spiking neural networks," Proceedings of NIPS-2011, 2011, https://proceedings.neurips.cc/paper/2011/file/795c7a7a5ec6b460ec00c5841019b9e9-Paper.pdf

[21] M. Awiszus, B. Rosenhahn, "Markov Chain Neural Networks," Proceeding of IEEE Conference on Computer Vision and Pattern Recognition Workshop on Vision With Biased or Scarce Data, 2018, https://openaccess.thecvf.com/content_cvpr_2018_workshops/papers/w42/Awiszus_Markov_Chain_Neural_CVPR_2018_paper.pdf

[22] P.A. Merolla, J. V. Arthur, R. Alvarez-Icaza, A. S. Cassidy, J. Sawada, F. Akopyan, F. et al., "A million spiking-neuron integrated circuit with a scalable communication network and interface" Science, 345, 6197, pp 668-673, 2014.

[23] C. Huang, A. Resnik, T. Celikel and B. Englitz, "Adaptive spike threshold enables robust and temporally precise neuronal encoding," PLoS computational biology, 12, 6, e1004984, 2016.

[24] M. Kiselev, "Rate coding vs. temporal coding-is optimum between?" Proceedings of 2016 international joint conference on neural networks (IJCNN), pp 1355-1359, 2016.

[25] M. Kiselev, "A synaptic plasticity rule providing a unified approach to supervised and unsupervised learning," Proceedings of 2017 International Joint Conference on Neural Networks (IJCNN), pp 3806-3813, 2017.

[26] M. Kiselev, and A. Lavrentyev, "A preprocessing layer in spiking neural networks–structure, parameters, performance criteria," Proceedings of 2019 International Joint Conference on Neural Networks (IJCNN), paper N-19450, 2019.